\documentclass{llncs}

\usepackage[year=2026]{eccv}

\usepackage{eccvabbrv}

\usepackage{graphicx}
\usepackage{booktabs}
\usepackage{caption}
\usepackage[accsupp]{axessibility}
\usepackage{mathtools}
\usepackage{amsmath}
\usepackage{amssymb}
\usepackage{multirow}
\usepackage{rotating}
\usepackage{adjustbox}
\usepackage{array}
\usepackage[hyphens]{url}

\usepackage{hyperref}
\usepackage{orcidlink}

\title{Efficient Video Diffusion with Sparse Information Transmission for Video Compression}

\author{
    Mingde Zhou$^{1}$\enspace
    Zheng Chen$^{1}$\enspace 
    Yulun Zhang$^{1}$\thanks{Corresponding author: Yulun Zhang, yulun100@gmail.com}
}
\institute{
  Shanghai Jiao Tong University
}

\begin{document}
\maketitle
\pagestyle{plain}

\begin{abstract}
Video compression aims to maximize reconstruction quality with minimal bitrates. Beyond standard distortion metrics, perceptual quality and temporal consistency are also critical. However, at ultra-low bitrates, traditional end-to-end compression models tend to produce blurry images of poor perceptual quality. Besides, existing generative compression methods often treat video frames independently and show limitations in time coherence and efficiency. To address these challenges, we propose the Efficient Video \textbf{Diff}usion with \textbf{S}parse \textbf{I}nformation \textbf{T}ransmission (Diff-SIT), which comprises the Sparse Temporal Encoding Module (STEM) and the One-Step Video Diffusion with Frame Type Embedder (ODFTE). The STEM sparsely encodes the original frame sequence into an information-rich intermediate sequence, achieving significant bitrate savings. Subsequently, the ODFTE processes this intermediate sequence as a whole, which exploits the temporal correlation. During this process, our proposed Frame Type Embedder (FTE) guides the diffusion model to perform adaptive reconstruction according to different frame types to optimize the overall quality. Extensive experiments on multiple datasets demonstrate that Diff-SIT establishes a new state-of-the-art in perceptual quality and temporal consistency, particularly in the challenging ultra-low-bitrate regime. Code is released at \url{https://github.com/MingdeZhou/Diff-SIT}.
\end{abstract}

\vspace{-2.5em}

\section{Introduction}
Video compression refers to encoding digital video streams into compact bitstreams, which facilitates efficient storage and transmission over bandwidth-limited channels. Generally, raw video data possesses immense volume and high redundancy, necessitating compression algorithms to exploit spatial correlations within frames and temporal dependencies across frames. Video compression has become the fundamental infrastructure for numerous multimedia and computer vision tasks, \emph{e.g.}, internet video streaming~\cite{sodagar2011mpeg, pantos2017http, lederer2013dynamic, ozbek2019implementation}, real-time video conferencing~\cite{jennings2023webrtc, min2025multi, lee2023r}, \emph{etc.} The primary challenge in this field lies in the rate-distortion optimization, which aims to reconstruct high-quality visual content using the minimum possible number of bits. To address this, conventional standards like H.265/HEVC~\cite{sullivan2012overview, ohm2012comparison, correa2021analysis, alariao2019sum} and H.266/VVC~\cite{bross2021overview, wang2021high, luo2023comparative, diaz2021vvc} utilize hand-crafted modules for block-based residual coding, prediction and transform. Recently, researchers have turned to neural video compression (NVC) frameworks. These end-to-end learning-based methods optimize the entire compression pipeline globally~\cite{Lu2019, li2021deep, balle2018variational, wu2018video}. Among these methods, the DCVC series~\cite{li2021deep, li2023neural, li2022hybrid, li2024neural} established a new paradigm by replacing residual coding with conditional probability modeling, setting a new benchmark for NVC performance.

\begin{figure}[t]
\centering
\includegraphics[width=\linewidth]{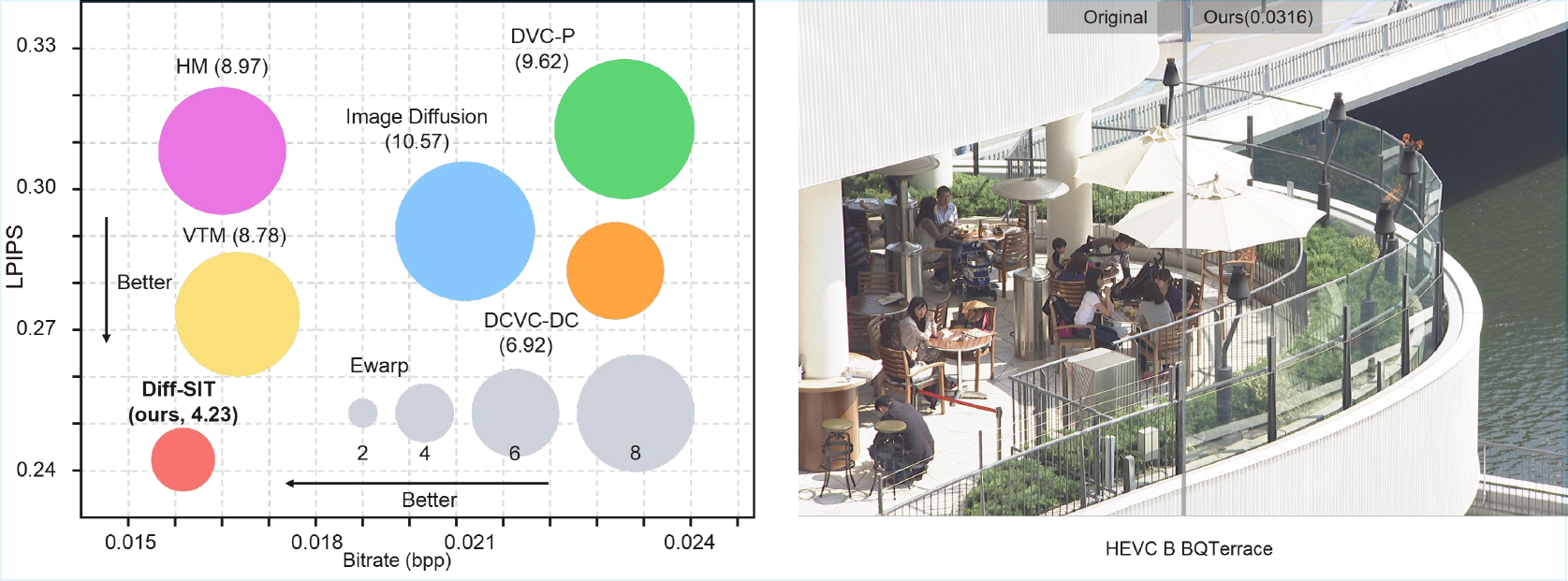}
\caption{(1) LPIPS-bitrate-coherence comparison on MCL-JCV. Temporal coherence is measured by Ewarp~\cite{lai2018learning}. Our method achieves the best perceptual quality as well as much higher temporal coherence. Because all the diffusion-based video compression methods have not opened source, we choose diffusion-based image compression method (SODEC)~\cite{chen2025steering} for comparison. (2) Qualitative comparison on HEVC class B BQTerrace. Our method recovers fine-grained textures with high perceptual quality.}
\label{fig:introduction}
\end{figure}

Recently, it has been recognized that optimizing for pixel-wise distortion metrics like PSNR is sub-optimal in the ultra-low bitrate regime. Strictly pursuing high PSNR often compels models to produce over-smoothed and blurry artifacts, failing to align with human perceptual realism~\cite{blau2018perception, wang2004image, johnson2016perceptual}. Consequently, researches have shifted towards perceptual quality-oriented video compression, aiming to synthesize rich, realistic textures even under extreme bandwidth constraints~\cite{santurkar2018generative, rippel2021improved, ledig2017photo}. Representative paradigms, such as GAN-based~\cite{ledig2017photo, mentzer2020high, blau2018perception} and diffusion-based models~\cite{dhariwal2021diffusion, guo2024low, kim2024diffusion}, employ generative reconstruction to synthesize missing details. However, these methods face critical limitations. Many generative approaches process video frames independently, neglecting vital spatiotemporal correlations~\cite{xie2021temporal, mirakhorli2023temporal}. For example, some diffusion compression methods perform image diffusion frame by frame, which leads to temporal inconsistency. Besides, most diffusion-based solutions rely on iterative multi-step sampling, introducing prohibitive latency that hinders real-world application~\cite{song2020denoising, kim2024diffusion, kim2024fast}.

To address these gaps, we propose the Efficient Video \textbf{Diff}usion with \textbf{S}parse \textbf{I}nformation \textbf{T}ransmission (Diff-SIT). \textbf{Firstly}, we posit that a true video diffusion model, which processes sequences holistically, is required to fully exploit temporal correlations and enhance temporal coherence. Thus, we propose the One-Step Video Diffusion with Frame Type Embedder (ODFTE) to efficiently perform diffusion to a whole frame sequence. Specifically, the diffusion process is guided by the Frame Type Embedder (FTE) to perform adaptive reconstruction to achieve better perception quality. \textbf{Secondly}, we argue that the powerful generative capability of video diffusion makes fully encoding every frame unnecessary. Based on this insight, we propose the Sparse Temporal Encoding Module (STEM) for sparse information transmission. Instead of fully compressing every frame, STEM only fully encodes a sparse set of backbone frames. Then, it represents all the other frames (we call them MV frames) using only low-bitrate optical flow fields. The ODFTE module is then employed to generatively reconstruct the final high-quality frame sequence. Here, the FTE refers to the frame type (backbone frame or MV frame) to guide the diffusion process.

Our main contributions are summarized as follows:
\begin{itemize}
    \item We present a one-step video diffusion model, which to the best of our knowledge is the first attempt at one-step video diffusion for this task.

    \item We design a novel Sparse Temporal Encoding Module (STEM) that significantly saves bitrate via sparse backbone and MV compression.

    \item We propose the Frame Type Embedder to guide the one-step diffusion process for higher reconstruction quality.

    \item Our framework achieves SOTA rate-perception performance on multiple datasets, while demonstrating massive advancement in temporal coherence.
\end{itemize}

\section{Related Works}
\subsection{Learned and Generative Video Compression}

Traditional video compression codecs, such as HEVC (HM) and VVC (VTM)~\cite{bross2021overview}, define the state-of-the-art in distortion metrics (e.g., PSNR) but yield poor perceptual quality at ultra-low bitrates. As an alternative, end-to-end Neural Video Codecs (NVCs) evolved from early residual-based methods~\cite{Lu2019,Agustsson2020ScaleSpace,Lin2020MLVC} to more powerful conditional coding frameworks~\cite{Habibian2019Autoencoder,Liu2020ConditionalEntropy,Ladune2021ConditionalFlexible}. Notably, the DCVC series~\cite{li2021deep,Sheng2023TCM,li2023neural} has achieved SOTA rate-distortion performance by leveraging deep contextual information. Concurrent to this, generative video compression methods~\cite{Kim2020Adversarial,Mentzer2022GANSynthesis,Li2024ExtremeDiffusion} have focused on maximizing perceptual quality, often at the expense of traditional metrics. These models, such as DVC-P~\cite{zhang2021dvcp,yang2022perceptual}, utilize strong generative priors to synthesize realistic details, proving highly effective in the challenging low-bitrate regime where traditional codecs fail.

\subsection{Diffusion Models in Video Compression}
Diffusion models~\cite{ho2020denoising,song2021scorebased,sohl2015deep,chen2025steering} have shown state-of-the-art performance in image generation, which has inspired their application in perceptual image compression~\cite{mentzer2020high,hoogeboom2023high,yang2023lossy,xia2025diffpc}. Concurrently, dedicated video diffusion models have been developed to explicitly model the temporal dimension, enabling the generation of coherent video sequences~\cite{ho2022video,ho2022imagen,singer2022make}. This potential has inspired applications in video compression. These methods, however, face two critical limitations. First, approaches like DiffVC~\cite{ma2025diffusion} apply image diffusion models frame-by-frame. This independent generative process ignores temporal correlations and may introduce temporal incoherence. Second, existing diffusion-based methods like GiViC and DiffVC~\cite{gao2025givic,ma2025diffusion} rely on multi-step diffusion process, which incurs a substantial computational cost and limit their applications~\cite{salimans2022progressive,song2020denoising}.

\subsection{Optical Flow in Video Compression}
Optical flow, which estimates per-pixel motion to exploit temporal redundancy, is fundamental to video compression~\cite{Jain1979}. Modern flow estimation is dominated by deep networks like PWC-Net~\cite{Sun2018} and RAFT~\cite{Teed2020}, which replaced classic optimization methods~\cite{Horn1981,Lucas1981}. In Neural Video Compression (NVC), flow is central to motion compensation. Early works like DVC~\cite{Lu2019} used it to warp pixels and compress the residual~\cite{Lu2019,Djelouah2019}. More advanced conditional codecs, such as the DCVC series~\cite{li2021deep}, warp features in the latent space to provide temporal context for the encoder and entropy model~\cite{Sheng2022,Li2022,li2023neural}. In these frameworks, the motion vector (MV) field itself must be compressed and transmitted as side information~\cite{Qi2023}. However, recursive prediction in NVC often suffers from error accumulation and quality collapse over long sequences~\cite{Lu2020,Li2024}. Studies investigating this trade-off~\cite{Lu2020} highlight that while short-term flow prediction achieves high accuracy, long-term propagation necessitates complex correction mechanisms~\cite{Li2024,Jiang2025}.

\section{Methodology}
\begin{figure*}[t]
    \centering
    \includegraphics[width=\linewidth]{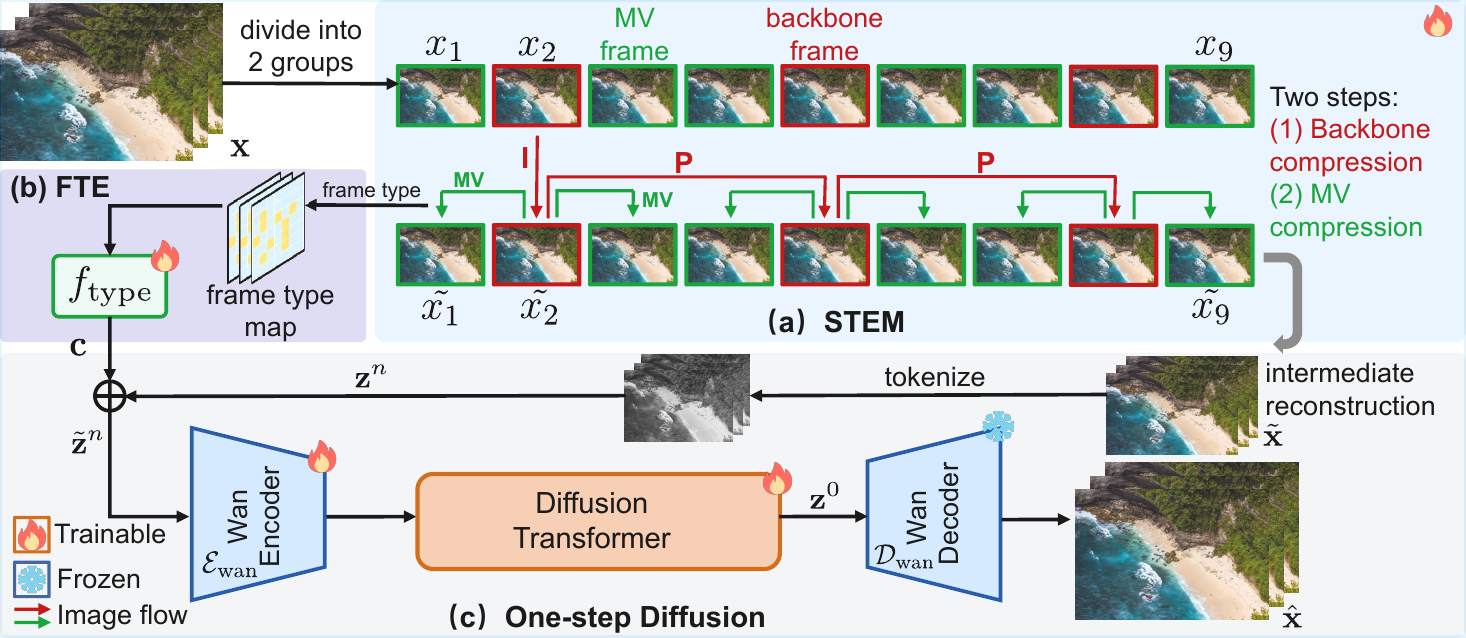}
    \caption{The overall architecture of our proposed Diff-SIT model. It consists of two main part: the STEM and ODFTE. Given an input frame sequence $\mathbf{x}$, first, it will be divided into backbone frames and MV frames. In STEM, there are two steps: \textbf{(1) Backbone compression:} the backbone frames are first compressed and reconstructed. \textbf{(2) MV compression:} each reconstructed backbone frame is used as a reference and MV compression is conducted to compress all the MV frames. After the STEM, the intermediate reconstructed sequence $\tilde{\mathbf{x}}$ will be fed into the ODFTE to be restored into final frame sequence $\hat{\mathbf{x}}$. Image flow refers to the coding order of video frames.}
    \label{fig:overview}
\end{figure*}

In this section, we provide an overview of our model, Efficient Video Diffusion with Sparse Information Transmission (Diff-SIT). Fig.~\ref{fig:overview} illustrates the overall architecture. The model is divided into two modules: the Sparse Temporal Encoding Module (STEM) and the One-Step Video Diffusion with Frame Type Embedder (ODFTE). The ODFTE can also be split into two parts: the Frame Type Embedder (FTE) and the One-step Video Diffusion.

Given an input frame sequence $\mathbf{x} = \{x_1, \dots, x_T\}$, we first select frames with indexes $3t+2$ as backbone frames and group them into a new backbone sequence $\{x_2, x_5, x_8, \dots\}$. The remaining frames are designated as Motion Vector (MV) frames. \textbf{Backbone compression:} The Sparse Temporal Encoding Module (STEM) first compresses and reconstructs the backbone sequence, yielding intermediate reconstructed backbone frames $\{\tilde{x}_2, \tilde{x}_5, \tilde{x}_8, \dots\}$. \textbf{MV compression:} STEM then proceeds to compress the MV frames. For each intermediate reconstructed backbone frame $\tilde{x}_{3t+2}$, the model estimates optical flows to its two neighboring original frames $x_{3t+1}$ and $x_{3t+3}$ and transmits the compressed flow fields. At decoder, decoded flows are used to warp the backbone frame, and the warped results are directly taken as the intermediate reconstructed MV frames,
\begin{equation}
    \label{eq:warping}
    \bigl[\tilde{x}_{3t+1},\, \tilde{x}_{3t+3}\bigr]
    = \mathcal{W}\Bigl(\tilde{x}_{3t+2},\,
    \bigl[\hat{\mathbf{F}}_{\tilde{x}_{3t+2} \to x_{3t+1}},\,
          \hat{\mathbf{F}}_{\tilde{x}_{3t+2} \to x_{3t+3}}\bigr]\Bigr).
\end{equation}
where $\tilde{x}_{3t+1}$ and $\tilde{x}_{3t+3}$ denote the reconstructed MV frames and $\hat{\mathbf{F}}_{\tilde{x}_{3t+2} \to x_{3t+1}}$, $\hat{\mathbf{F}}_{\tilde{x}_{3t+2} \to x_{3t+3}}$ are the optical flows from the reconstructed backbone frame $\tilde{x}_{3t+2}$ to the original neighboring frames $x_{3t+1}$ and $x_{3t+3}$, respectively. Here, $\mathcal{W}$ denotes the warping operator, which warps frame $\tilde{x}_{3t+2}$ via optical flows $\mathbf{\hat{F}}$. Finally, the intermediate reconstructed backbone frames and MV frames are rearranged into the original order to form the full intermediate reconstructed sequence $\tilde{\mathbf{x}}$.

The full intermediate sequence $\tilde{\mathbf{x}}$ is then fed into the ODFTE to be restored into a high-quality frame sequence $\hat{\mathbf{x}}$. \textbf{Frame Type Embedder (FTE):} First, FTE generates a type embedding $\mathbf{c}$ based on the compression type of each frame (backbone or MV). \textbf{One-step Video Diffusion:} Then, conditioned on this type embedding $\mathbf{c}$, the diffusion model performs adaptive diffusion reconstruction. Thus, we gain the final reconstructed sequence $\hat{\mathbf{x}}$.

\subsection{Sparse Temporal Encoding Module (STEM)}

In our STEM, we first perform the \textbf{backbone compression}, where the backbone frames are compressed as an independent video sequence. \textbf{The I-frame compression:} For the first backbone frame $x_{2}$ ($x_{3t+2}$ where $t=0$), we perform an independent intra-coding (I-frame encoding). Specifically, we use a pre-trained image compression model (HiFiC)~\cite{mentzer2020high} for I-frame encoding. All subsequent backbone frames are predictively coded as P-frames, referencing the previously reconstructed frame. The I-frame and P-frames are both the backbone frames.

\begin{figure}[t]
    \centering
    \includegraphics[width=\linewidth]{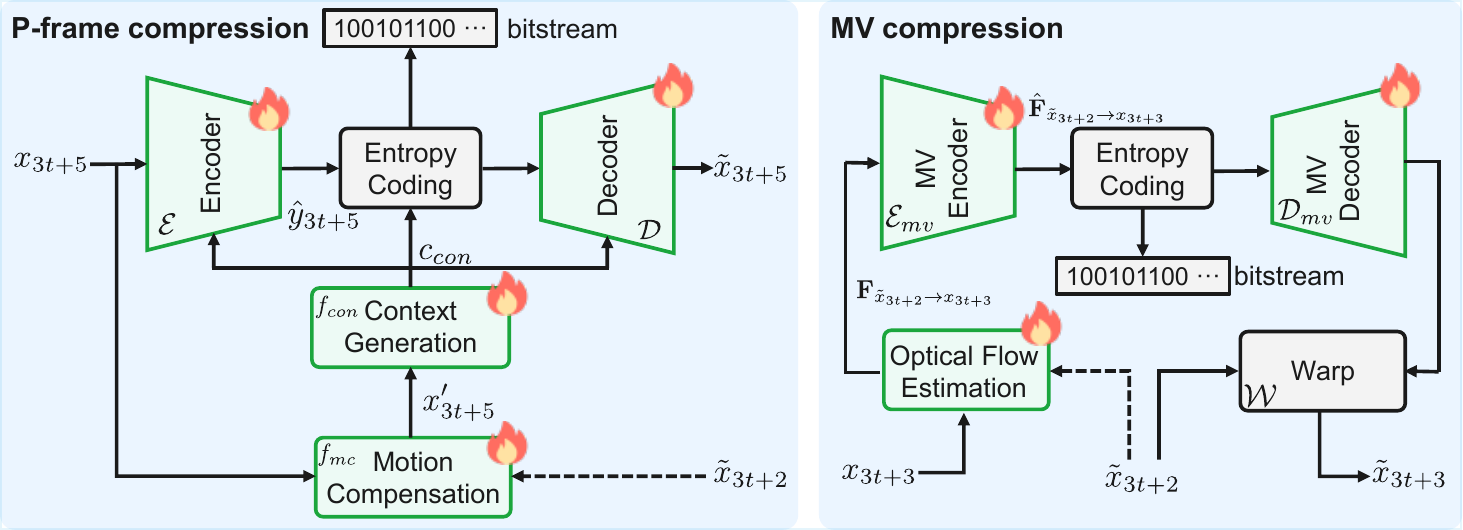}
    \caption{Pipeline of P-frame compression and MV compression. In \textbf{P-frame compression}, the reference frame $\tilde{x}_{3t+2}$ is used to compress the target frame $x_{3t+5}$ via conditional encoding. In \textbf{MV compression}, the reference frame $\tilde{x}_{3t+2}$ is used to compress frames $x_{3t+1}$ and $x_{3t+3}$ via MV compression. The reconstructed flow field $\hat{\mathbf{F}}_{\tilde{x}_{3t+2} \to x_{3t+1}}$ and $\hat{\mathbf{F}}_{\tilde{x}_{3t+2} \to x_{3t+3}}$ are used to warp the reference frame $\tilde{x}_{3t+2}$ to reconstruct the MV frames ($\tilde{x}_{3t+1}$) and ($\tilde{x}_{3t+3}$) respectively. Here, we use $\tilde{x}_{3t+3}$ as an example.}
    \label{fig:STEM}
\end{figure}

\textbf{Details of P-frame compression:} As shown in Fig.~\ref{fig:STEM}, the pipeline of P-frame compression is:
\textbf{(1) Motion compensation:} 
We perform a motion compensation, which consumes a few bits. We choose SpyNet~\cite{ranjan2017optical} as our optical flow estimation network. SpyNet is used to produce the motion-compensated estimation ($x'_{3t+5}$) of the original frame $x_{3t+5}$,
\begin{equation}
    x'_{3t+5} = f_{mc} (\tilde{x}_{3t+2}, x_{3t+5}).
\end{equation}
where $f_{mc}$ is the motion compensation network.
\textbf{(2) Conditional encoding:} Following the method proposed in DCVC~\cite{li2021deep,li2023neural}, we perform a conditional encoding framework. First, the context generation module ($f_{con}$) generates a context feature $c_{con}$ based on the motion-compensated estimation $x'_{3t+5}$ we get,
\begin{equation}
    c_{con} = f_{con}(x'_{3t+5}).
\end{equation}
Subsequently, the encoder $\mathcal{E}$ compresses the current frame $x_{3t+5}$ into a latent representation $y_{3t+5}$ conditioned on this context $c_{con}$. Then $y_{3t+5}$ is quantized, $\hat{y}_{3t+5} = \mathcal{Q}(y_{3t+5})$, where $\mathcal{Q}$ refers to the quantization operation. An entropy model estimates the probability distribution of quantized latent $\hat{y}_{3t+5}$ conditioned on the context $c_{con}$, and performs an entropy encoding to generate a bitstream. After the bitstream is transmitted, a decoder $\mathcal{D}$, again guided by $c_{con}$, reconstructs from the decoded latent representation $\hat{y}_{3t+5}$ to get the intermediate reconstructed frame $\tilde{x}_{3t+5}$. The overall pipeline can be shown as
\begin{equation}
\tilde{x}_{3t+5} = \mathcal{D}(\mathcal{Q}(\mathcal{E}(x_{3t+5},c_{con})), c_{con}).
\end{equation}
Performing this process on all backbone frames, and we can get the intermediate reconstructed backbone frames $\{\tilde{x}_2, \tilde{x}_5, \tilde{x}_8, \dots\}$.

Then, we proceed to compress the remaining MV frames, performing the \textbf{MV compression}. This process is performed in parallel for all MV frames. Each reconstructed backbone frame $\tilde{x}_{3t+2}$ serves as a reference for its two neighboring frames $x_{3t+1}$ and $x_{3t+3}$. The ``MV Compression'' pipeline is shown in Fig.~\ref{fig:STEM}.

Using $\tilde{x}_{3t+2}$ as the reference, an optical flow estimation~\cite{ranjan2017optical} module predicts the motion from $\tilde{x}_{3t+2}$ to $x_{3t+1}$ and $x_{3t+3}$ respectively, denoted as motion vector field $\mathbf{F}_{\tilde{x}_{3t+2} \to x_{3t+1}}$, $\mathbf{F}_{\tilde{x}_{3t+2} \to x_{3t+3}}$. Each field $\mathbf{F}_{\tilde{x}_{3t+2} \to x_{3t+5}}$. $\mathbf{F}_{\tilde{x}_{3t+2} \to x_{3t+5}}$ is then passed through an MV Encoder ($\mathcal{E}_{mv}$) to produce a compact latent representation, and then quantized and entropy-coded into a bitstream for transmission. This step consumes few bits, because we only need to transmit the motion vector field $\hat{\mathbf{F}}$. At the decoder, the bitstream is entropy-decoded and processed by an MV Decoder ($\mathcal{D}_{mv}$) to obtain the reconstructed flow field $\hat{\mathbf{F}}$. Using $\mathbf{F}_{\tilde{x}_{3t+2} \to x_{3t+3}}$ as an example, the pipeline for transmitting vector flow field can be shown as: 
\begin{equation}
    \hat{\mathbf{F}}_{\tilde{x}_{3t+2} \to x_{3t+3}} = \mathcal{D}_{mv}(\mathcal{Q}(\mathcal{E}_{mv}(\mathbf{F}_{\tilde{x}_{3t+2} \to x_{3t+3}})),
\end{equation}
where $\mathcal{Q}$ refers to the quantization operation. Specifically, for a given reconstructed backbone frame $\tilde{x}_{3t+2}$, it serves as the reference frame to compress both the preceding frame $x_{3t+1}$ and the succeeding frame $x_{3t+3}$ via MV compression. The reconstructed flow $\hat{\mathbf{F}}$ is used to warp the reference frame $\tilde{x}_{3t+2}$ to produce intermediate reconstructed MV frames $\tilde{x}_{3t+1}$ and $\tilde{x}_{3t+3}$.
\begin{equation}
    \label{eq:warping2}
    \bigl[\tilde{x}_{3t+1},\, \tilde{x}_{3t+3}\bigr]
    = \mathcal{W}\Bigl(\tilde{x}_{3t+2},\,
    \bigl[\hat{\mathbf{F}}_{\tilde{x}_{3t+2} \to x_{3t+1}},\,
          \hat{\mathbf{F}}_{\tilde{x}_{3t+2} \to x_{3t+3}}\bigr]\Bigr).
\end{equation}
To be more clear, the intermediate reconstructed MV frames $\tilde{x}_{3t+1}$ and $\tilde{x}_{3t+3}$ are exactly the warped results of the reference backbone frame $\tilde{x}_{3t+2}$. In this manner, we reconstruct all backbone and MV frames, yielding the full intermediate frame sequence $\tilde{\mathbf{x}}$$=$$\{\tilde{x}_1, \dots, \tilde{x}_T\}$ for later diffusion reconstruction.

\begin{figure}[t]
    \centering 
    \includegraphics[width=\linewidth]{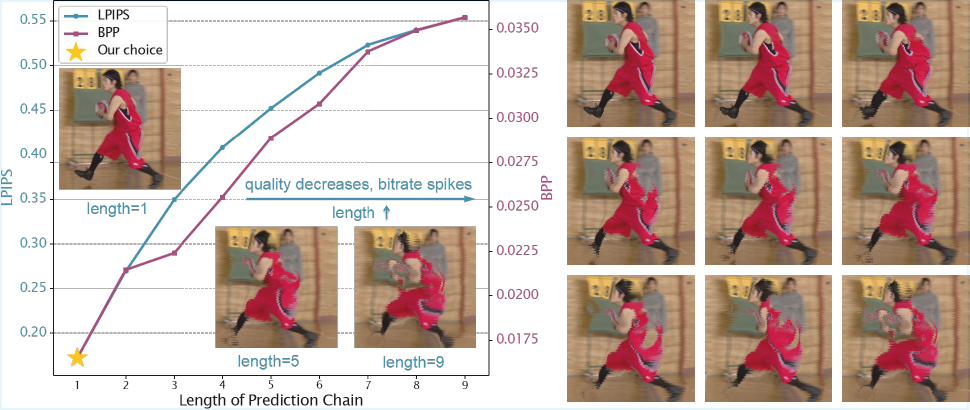}
    \caption{Analysis of reconstruction quality and bitrate cost versus the length of the continuous optical flow prediction chain. Reconstruction quality rapidly degrades as the prediction chain grows, while the required bitrate also increases significantly. Image at right is the corresponding 1-9 frames, from top-left to bottom-right.}
    \label{fig:flow_length}
\end{figure}

The essence of MV compression is an optical flow compression. While many studies have discussed the problem of quality collapse in long optical flow prediction chains, we find that for very short chains, flow-only compression can achieve superior reconstruction results at a lower bitrate. As shown in Fig.~\ref{fig:flow_length}, the reconstruction quality rapidly degrades as optical prediction chains become longer. Thus, we choose to limit the optical prediction length to only one immediate neighbor (a reference frame $\tilde{x}_{3t+2}$ only serves for MV-compressing $x_{3t+1}$ and $x_{3t+3}$), which achieves excellent reconstruction quality at a minimal bit cost.

\subsection{One-Step Video Diffusion with Frame Type Embedder (ODFTE)}
We introduce the ODFTE module to perform generative reconstruction by leveraging the powerful capabilities of diffusion models to reinforce the intermediate reconstructed sequence $\tilde{\mathbf{x}}$. This module consumes zero bits.

\textbf{Frame Type Embedder (FTE):} First, we use the FTE to extract a type-bias embedding $\mathbf{c}$ from the encoding type of the frame sequence. For instance, with $T=9$, the encoding type sequence is [MV, I, MV, MV, P, MV, MV, P, MV]. We first assign a unique one-hot vector to each type (I, P, or MV) and broadcast this vector to the full spatial dimensions of the frame. These broadcasted maps are then concatenated along the temporal dimension to create the frame type map $\mathbf{M}$, as shown in Fig.~\ref{fig:overview}. Subsequently, we apply a lightweight 3D convolution network $f_{type}$ to this map to extract the type information, which is then flattened to produce the final type-bias embedding $\mathbf{c}$, i.e., $\mathbf{c}=f_{type}(\mathbf{M})$.

\textbf{One-step Video Diffusion:} Concurrently, we utilize the pre-trained Diffusion Transformer (DiT)~\cite{peebles2023scalable} Wan 2.1 model~\cite{wan2025} for denoising. First, the Wan Encoder $\mathcal{E}_{wan}$ encodes the intermediate frame sequence $\tilde{x}$ into a latent representation. Subsequently, the tokenizer converts this representation into a token sequence $\mathbf{z}^n$, where $n$ denotes the diffusion timestep.
\textbf{To maximize efficiency,} we fix the timestep $n$ and employ a one-step diffusion process. This is motivated by our observation that the input $\tilde{\mathbf{x}}$ already contains rich information from the STEM, meaning that a single refinement step is sufficient to restore fine-grained details, obviating the need for an iterative denoising process from pure noise. 
We also apply \textbf{adaptive diffusion strength}. Different frames in the intermediate frame sequence $\tilde{\mathbf{x}}$ have different levels of quality (backbone frames v.s MV frames). We propose that one-step diffusion process should apply adaptive denoising, stronger to low-quality MV frames, while weaker to more reliable backbone frames, which can better preserve the original information. So, the type-bias embedding $\mathbf{c}$ is added to the tokens $z^n$ to obtain the type-biased tokens $\tilde{\mathbf{z}}^n$, where $\tilde{\mathbf{z}}^n$$=$$\mathbf{z}^n \oplus \mathbf{c}$. Thus, the diffusion model is informed and can condition on the type embedding $\mathbf{c}$ for adaptive reconstruction. Then, the one-step diffusion transformer denoises $\tilde{\mathbf{z}}^n$, yielding the clean tokens $\mathbf{z}^0$,
\begin{equation}
\label{eq:onestep_denoise}
\mathbf{z}^0 = \frac{1}{\sqrt{\bar{\alpha}_n}} \left( \tilde{\mathbf{z}}^n - \sqrt{1-\bar{\alpha}_n} \cdot \epsilon_\theta(\tilde{\mathbf{z}}^n, n) \right),
\end{equation}
where $\bar{\alpha}_n$ is the cumulative noise schedule parameter at the fixed timestep $n$, and $\epsilon_\theta$ is the noise-prediction network (DiT) parameterized by $\theta$.
Finally, $\tilde{\mathbf{z}}^0$ is transformed back into a latent representation and decoded by the 3D Wan Decoder $\mathcal{D}_{wan}$~\cite{wan2025} to produce the final reconstructed frame sequence $\hat{\mathbf{x}}$. Our ODFTE is designed on common DiT, other DiT backbones can also be applied.

\subsection{Training and Inferencing}
Our model is trained in two stages. 
In the first stage, we train only the backbone reconstruction part of the STEM. Sequence length $T$ is set to 9. Only backbone frames, i.e, sequence $\mathbf{x}_{3t+2}$$=$$\{x_2, x_5, x_8\}$ are used for training. STEM is optimized for the intermediate sequence $\tilde{\mathbf{x}}_{3t+2}$$=$$\{\tilde{x}_2, \tilde{x}_5, \tilde{x}_8\}$. The loss function is:
\begin{equation}
\mathcal{L}_{\text{s1}} = \lambda \cdot r(\hat{\mathbf{y}}_{3t+2}, f_{mc}) + \text{MSE}(\mathbf{x}_{3t+2}, \tilde{\mathbf{x}}_{3t+2}),
\label{eq:LossStage1}
\end{equation}
where $r(\cdot)$ denotes the bitrate cost. $\hat{\mathbf{y}}_{3t+2}$ is bitrate for the quantized backbone latents $\hat{\mathbf{y}}_{3t+2}$, and $f_{mc}$ means bitrate comsumed during the motion compensation. $\text{MSE}$ represents the Mean Squared Error. Hyperparameter $\lambda$ controls the rate-distortion trade-off. We load the pre-trained HiFiC model~\cite{mentzer2020high} for I-frame compression, and the P-frame compression module is trained end-to-end.

In the second stage, we load the pre-trained Wan 2.1 1.3B model~\cite{wan2025} and train the entire model (STEM and ODFTE) end-to-end. We still set $T$$=$$9$, but now use the full sequence $\mathbf{x}$$=$$\{x_1, \dots, x_9\}$ and optimize for the final high-quality reconstruction $\hat{\mathbf{x}}$$=$$\{\hat{x}_1, \dots, \hat{x}_9\}$. The loss function is:
\begin{equation}
    \label{eq:LossStage2}
    \begin{aligned}
    \mathcal{L}_{\text{s2}} =
    &\ \lambda \cdot r\bigl(\hat{\mathbf{y}}_{3t+2}, f_{mc}, \hat{\mathbf{F}}_{3t+2 \to 3t+1, 3t+3}\bigr)
    + \text{LPIPS}(\mathbf{x}, \hat{\mathbf{x}}) \\
    &\ + k_1 \cdot \text{MSE}(\mathbf{x}, \hat{\mathbf{x}})
    + k_2 \cdot \mathcal{L}_{\text{temp}}(\mathbf{x}, \hat{\mathbf{x}}).
    \end{aligned}
\end{equation}
where the rate term $r(\cdot)$ now includes all the bit cost, including 
(1) backbone compression cost: ($\hat{\mathbf{y}}_{3t+2}$, $f_{mc}$),
(2) MV compression cost: ($\hat{\mathbf{F}}_{3t+2 \to 3t+2 \pm  1}$),
and $\mathcal{L}_{\text{temp}}$ represents the Frame Difference Loss. Here, we use MSE Loss:
\begin{equation}
    \mathcal{L}_{\text{temp}} = \mathbb{E} \left[ \sum_{t=2}^{T} \left\| (\hat{x}_t - \hat{x}_{t-1}) - (x_t - x_{t-1}) \right\|_2^2 \right].
\end{equation}
For inferencing, our model can operate on sequences of arbitrary length $T$. To align with real-world GOP (Group of Pictures) sizes which are often 32 or larger, we set $T$$=$$33$. This sequence length consists of 11 backbone frames (1 I-frame, 10 P-frames) and 22 MV frames. We choose $T$$=$$33$ as Wan model requires the sequence length to be $8n$$+$$1$. We experimented on different density of backbone frames, and Tab.~\ref{table: interval} shows that current setting (interval 2, i.e [MV, P, MV, MV, P, MV...]) is optimal. More detailed analysis can be found in ablation study.

\section{Experiments}

\subsection{Experimental Settings}
\noindent\textbf{Datasets.}
We employ a two-stage training strategy. First, the STEM backbone is optimized via Eq.~\eqref{eq:LossStage1} on frames 1, 4, and 7 of the Vimeo-9k dataset~\cite{xue2019video} for 700k steps. Subsequently, the entire framework is fine-tuned end-to-end minimizing Eq.~\eqref{eq:LossStage2} on the REDS dataset~\cite{nah2019ntire} using random $256 \times 256$ 9-frame patches for 15,000 steps. We benchmark our model Diff-SIT on three standard high-resolution test datasets: HEVC B~\cite{JCTVC-L1100}, MCL-JCV~\cite{wang2016mcl}, and UVG~\cite{mercat2020uvg}. During inference, we evaluate all sequences at their full resolution.

\noindent\textbf{Metrics.}
We employ LPIPS~\cite{zhang2018unreasonable} and DISTS~\cite{ding2020image} to evaluate reference perceptual quality. CLIPIQA~\cite{wang2023exploring} is utilized to assess no-reference visual realism. Temporal consistency is measured by Ewarp~\cite{lai2018learning} via optical flow warping error. Finally, compression rate is reported in bits per pixel (bpp). Beyond quantitative analysis, we also provide visual results for straight-forward comparison in Fig~\ref{fig:QualitativeResult}.

\noindent\textbf{Implementation Details.}
We adopt the DCVC-DC architecture~\cite{li2023neural} for backbone P-frame encoding and utilize the Wan 2.1 model~\cite{wan2025} with a fixed timestep $n$$=$$799$ for ODFTE. Training employs the default AdamW optimizer~\cite{loshchilov2019decoupled} with batch sizes of 8 in Stage 1 and 2 in Stage 2. Loss hyperparameters are set to $\lambda$$=$$5$$\cdot$$10^{-3}$ for Eq.~\eqref{eq:LossStage1}, and $\lambda$$\in$$\{0.5, 1.0, 2.0, 4.0\}$, $k_1$$=$$10$, $k_2$$=$$0.1$ for Eq.~\eqref{eq:LossStage2}. Experiments are conducted on one NVIDIA RTX A6000 GPU, and the training takes about two days. More details are provided in supplementary materials.

\subsection{Main Results}

\newcommand{\QuantFigLeftShift}{-0pt} 
\newcommand{\QuantFigLabelSize}{\scriptsize}
\newcommand{\QuantFigColGap}{0mm}
\begin{figure*}[t]
    \centering
    \hspace{\QuantFigLeftShift}
    \begin{tabular}{@{}c@{\hspace{\QuantFigColGap}}c@{}}
        \rotatebox{90}{{\QuantFigLabelSize \ \ HEVC Class B}} & 
        \includegraphics[width=0.99\textwidth]{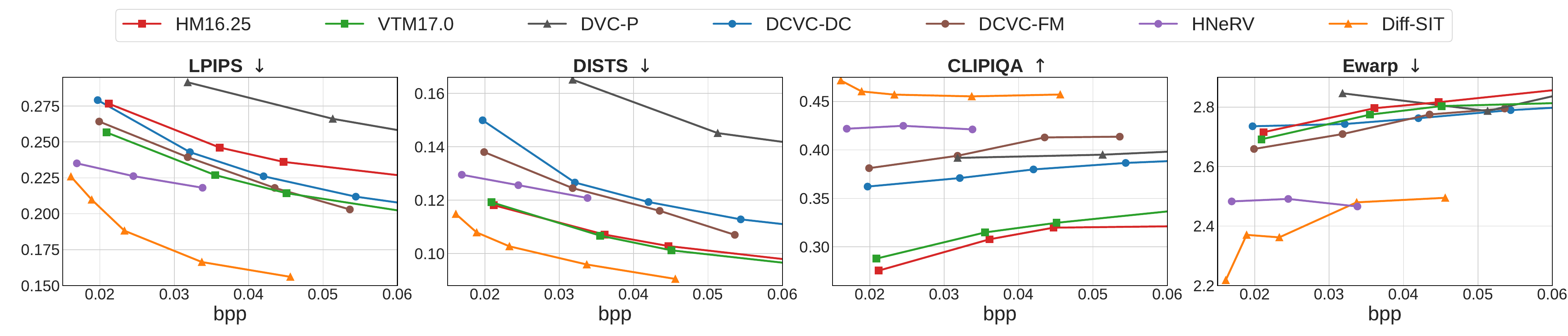} \\
        \addlinespace[-4pt]
        
        \rotatebox{90}{{\QuantFigLabelSize \ \ \ \ \ MCL-JCV}} & 
        \includegraphics[width=0.99\textwidth]{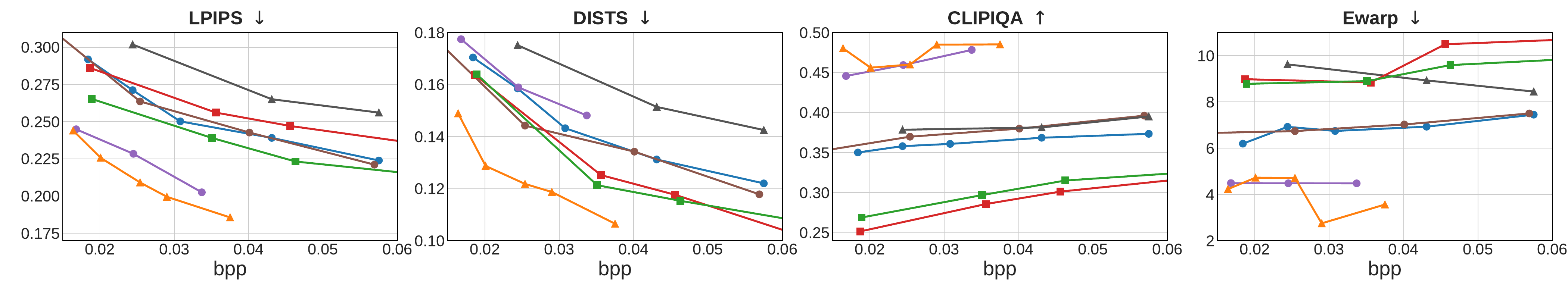} \\
        \addlinespace[-4pt]
        
        \rotatebox{90}{{\QuantFigLabelSize \ \ \ \ \ \ \ \ \ UVG}} & 
        \includegraphics[width=0.99\textwidth]{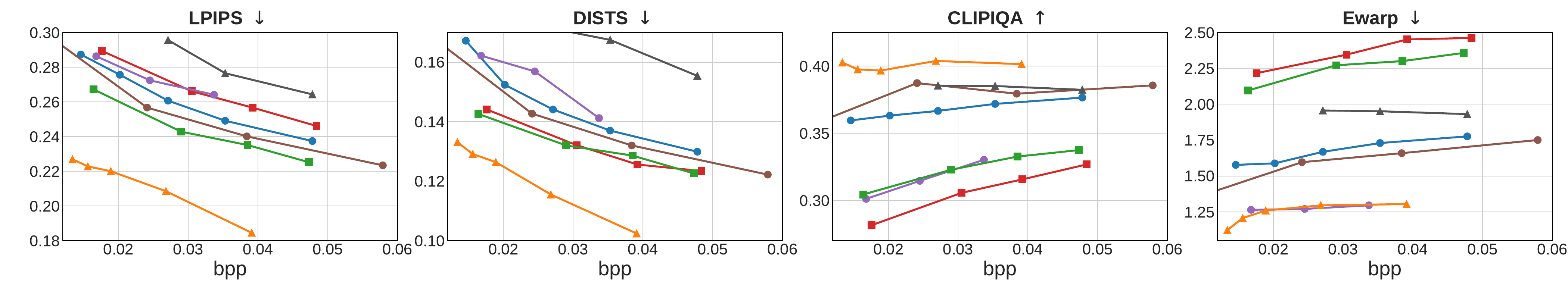}
    
    \end{tabular}
    \vspace{-2.mm}
    \caption{Quantitative comparison with state-of-the-art methods on the HEVC Class B, MCL-JCV and UVG datasets. $\downarrow$ means lower is better.}
    \label{fig:QuantitativeResult}
\end{figure*}

\newcommand{\figlabel}[1]{
    \makebox[\linewidth][c]{\QualThumbLabelSize \mbox{\textbf{#1}}}
}

\newcommand{\smallfig}[2]{
    \begin{tabular}{@{}p{\linewidth}@{}}
        \figlabel{#1}\\[2pt]
        \includegraphics[width=\linewidth]{#2}
    \end{tabular}
}

\newcommand{\QualLeftWidth}{0.157\linewidth}    
\newcommand{\QualThumbWidth}{0.18\linewidth}    
\newcommand{\QualRowVSpace}{0mm}                
\newcommand{\QualColSep}{2pt}                   
\newcommand{\QualLeftLabelSize}{\tiny}         
\newcommand{\QualThumbLabelSize}{\tiny}        
\newcommand{\QualLabelImageGap}{0.16\baselineskip} 
\newcommand{\QualLeftLabelVShift}{0.2pt}               

\newcommand{\qualblock}[3]{
  \begin{tabular}{@{}cc@{}}
    \begin{adjustbox}{valign=c}
      \begin{minipage}[c]{\QualLeftWidth}
        \centering
        \vspace{\QualLeftLabelVShift}
        {\QualLeftLabelSize \textbf{#1}}\\[\QualLabelImageGap]
        \includegraphics[width=\linewidth]{#2}
      \end{minipage}
    \end{adjustbox}
    &
    \begin{adjustbox}{valign=c}
      #3
    \end{adjustbox}
  \end{tabular}
}

\begin{figure*}[t]
    \centering
    \setlength{\tabcolsep}{\QualColSep}
    \renewcommand{\arraystretch}{0.1}

    \qualblock{HEVC Class B}{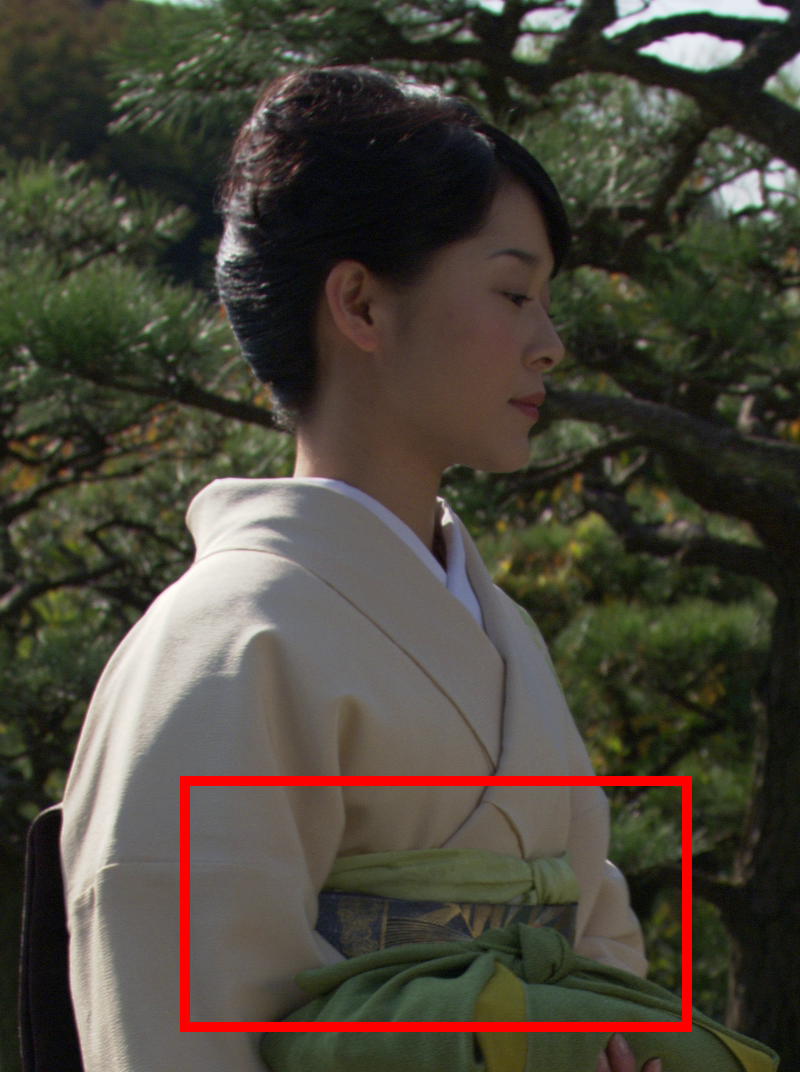}{
        \begin{tabular}{cccc}
            \begin{minipage}{\QualThumbWidth}\smallfig{GT (BPP$\downarrow$)}{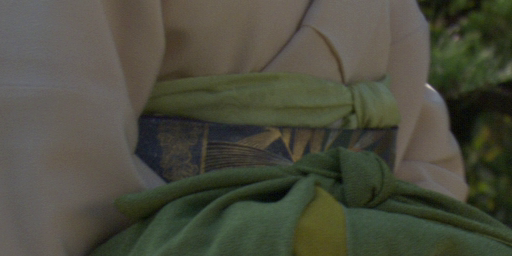}\end{minipage} &
            \begin{minipage}{\QualThumbWidth}\smallfig{HM (0.0451)}{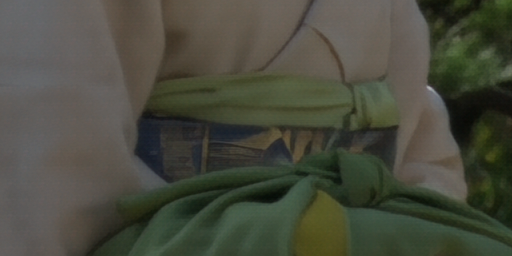}\end{minipage} &
            \begin{minipage}{\QualThumbWidth}\smallfig{VTM (0.0447)}{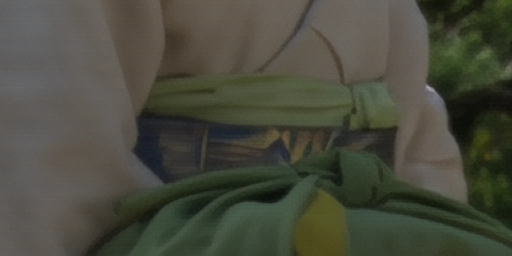}\end{minipage} &
            \begin{minipage}{\QualThumbWidth}\smallfig{DVC-P (0.0539)}{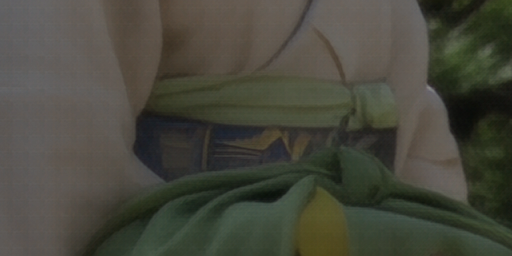}\end{minipage} \\
            \noalign{\vspace{4pt}}
            \begin{minipage}{\QualThumbWidth}\smallfig{DCVC-DC (0.0436)}{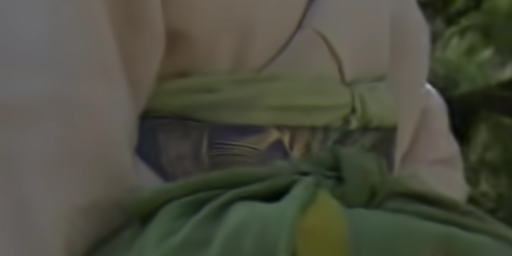}\end{minipage} &
            \begin{minipage}{\QualThumbWidth}\smallfig{DCVC-FM (0.0455)}{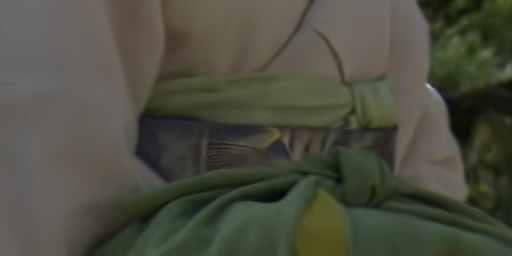}\end{minipage} &
            \begin{minipage}{\QualThumbWidth}\smallfig{HNeRV (0.0398)}{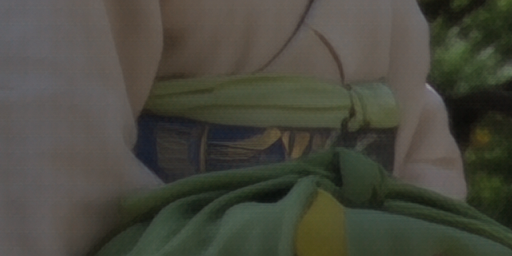}\end{minipage} &
            \begin{minipage}{\QualThumbWidth}\smallfig{Diff-SIT (0.0330)}{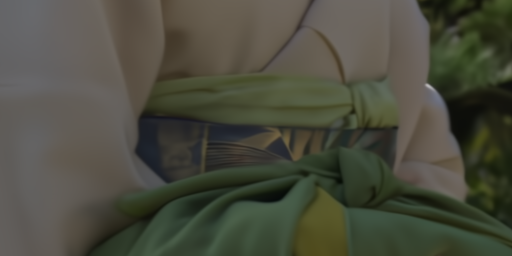}\end{minipage}
        \end{tabular}
    }

    \vspace{\QualRowVSpace}

    \qualblock{MCL-JCV}{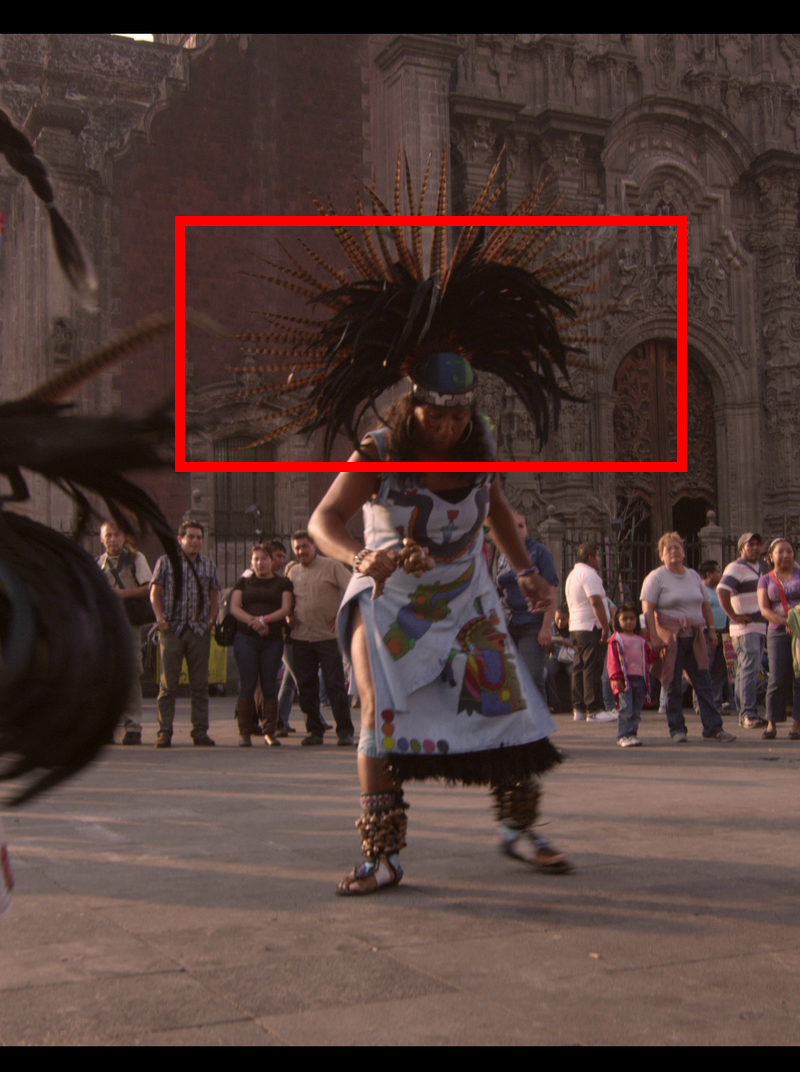}{
        \begin{tabular}{cccc}
            \begin{minipage}{\QualThumbWidth}\smallfig{GT (BPP$\downarrow$)}{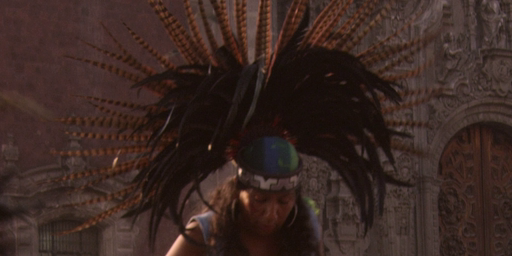}\end{minipage} &
            \begin{minipage}{\QualThumbWidth}\smallfig{HM (0.0392)}{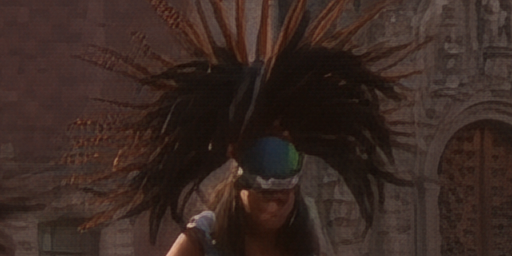}\end{minipage} &
            \begin{minipage}{\QualThumbWidth}\smallfig{VTM (0.0385)}{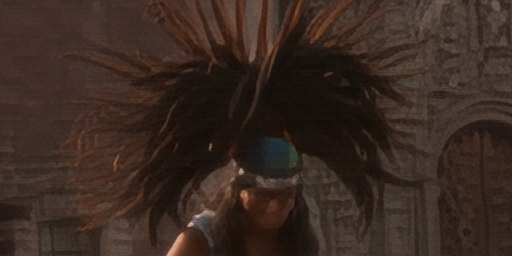}\end{minipage} &
            \begin{minipage}{\QualThumbWidth}\smallfig{DVC-P (0.0403)}{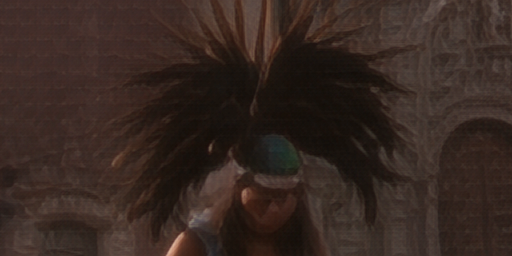}\end{minipage} \\
            \noalign{\vspace{4pt}}
            \begin{minipage}{\QualThumbWidth}\smallfig{DCVC-DC (0.0353)}{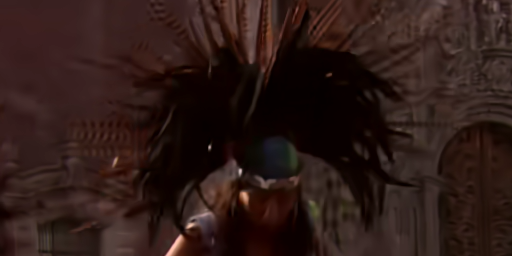}\end{minipage} &
            \begin{minipage}{\QualThumbWidth}\smallfig{DCVC-FM (0.0384)}{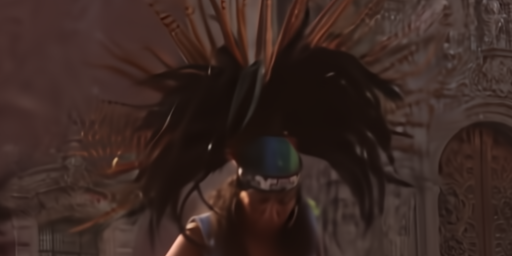}\end{minipage} &
            \begin{minipage}{\QualThumbWidth}\smallfig{HNeRV (0.0337)}{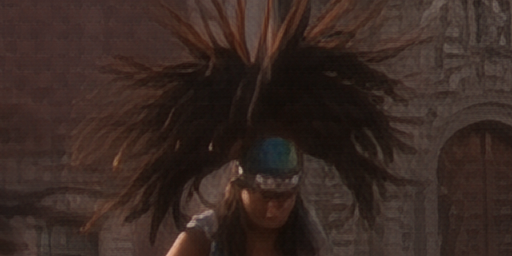}\end{minipage} &
            \begin{minipage}{\QualThumbWidth}\smallfig{Diff-SIT (0.0268)}{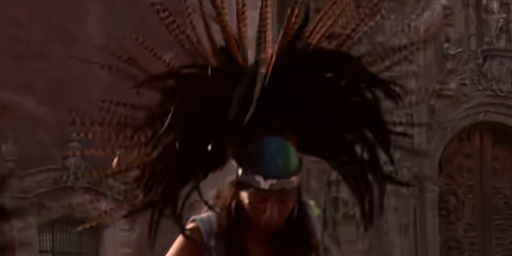}\end{minipage}
        \end{tabular}
    }

    \vspace{\QualRowVSpace}

    \qualblock{UVG}{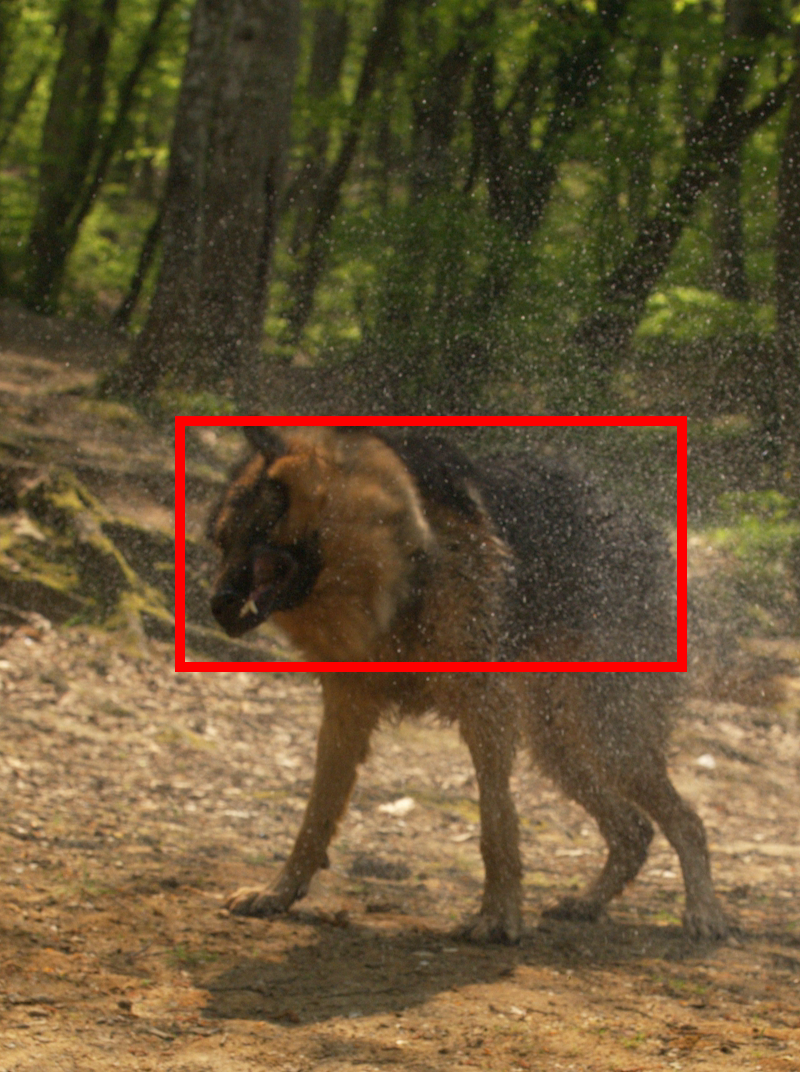}{
        \begin{tabular}{cccc}
            \begin{minipage}{\QualThumbWidth}\smallfig{GT (BPP$\downarrow$)}{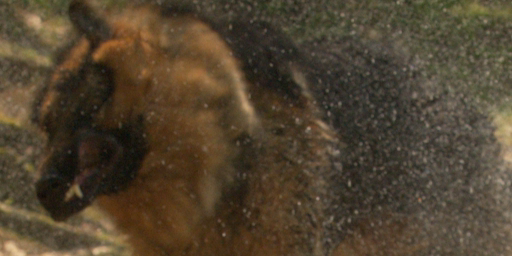}\end{minipage} &
            \begin{minipage}{\QualThumbWidth}\smallfig{HM (0.0352)}{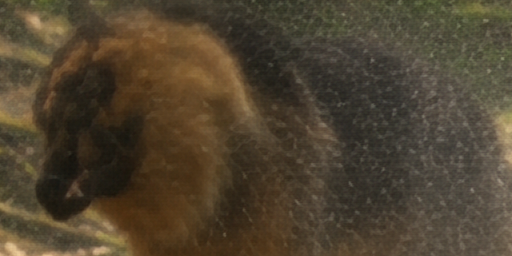}\end{minipage} &
            \begin{minipage}{\QualThumbWidth}\smallfig{VTM (0.0345)}{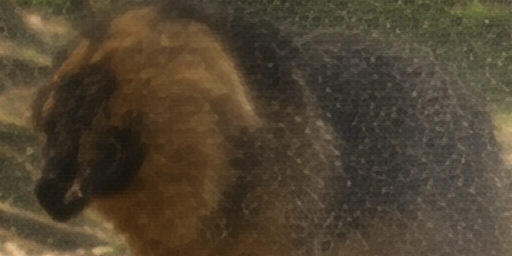}\end{minipage} &
            \begin{minipage}{\QualThumbWidth}\smallfig{DVC-P (0.0368)}{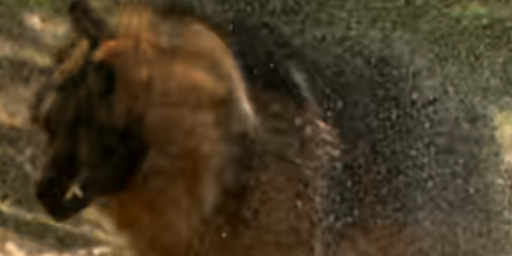}\end{minipage} \\
            \noalign{\vspace{4pt}}
            \begin{minipage}{\QualThumbWidth}\smallfig{DCVC-DC (0.0418)}{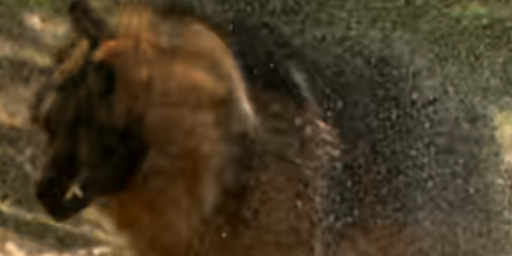}\end{minipage} &
            \begin{minipage}{\QualThumbWidth}\smallfig{DCVC-FM (0.0429)}{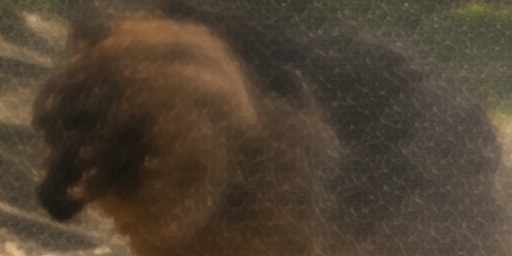}\end{minipage} &
            \begin{minipage}{\QualThumbWidth}\smallfig{HNeRV (0.0391)}{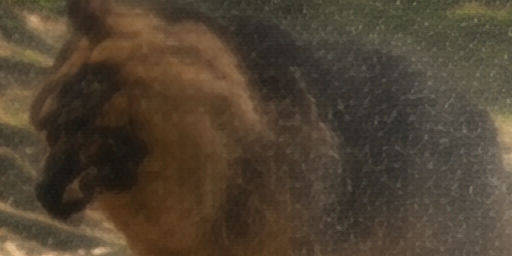}\end{minipage} &
            \begin{minipage}{\QualThumbWidth}\smallfig{Diff-SIT (0.0291)}{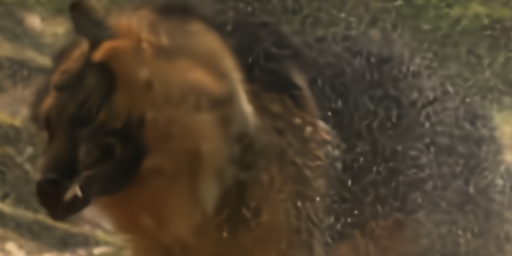}\end{minipage}
        \end{tabular}
    }

    \caption{Visual comparison on the HEVC Class B, MCL-JCV and UVG datasets.}
    \label{fig:QualitativeResult}
\end{figure*}

To rigorously validate the effectiveness of our model Diff-SIT, we benchmark against state-of-the-art methods spanning different paradigms. Specifically, we compare our approach with leading traditional standards, HM (H.265/HEVC)~\cite{sullivan2012overview} and VTM (H.266/VTM)~\cite{bross2021overview}, as well as representative end-to-end neural video compression (NVC) models such as DCVC-DC~\cite{li2023neural} and DCVC-FM~\cite{li2024dcvcfm}. To ensure a comprehensive evaluation against diverse strategies, we also include the GAN-based DVC-P~\cite{yang2022perceptual} and the implicit neural representation (INR) method HNeRV~\cite{chen2023hnerv}. As far as we know, until now, all existing diffusion-based methods such as GiViC~\cite{gao2025givic} and DiffVC~\cite{ma2025diffusion} haven't opened source.

\begin{table}[h]
    \centering
    \begin{minipage}[t]{0.48\linewidth}
        \centering
        \captionof{table}{Comparison on distortion metrics on HEVC class B. We report the PSNR and MS-SSIM at similar bitrates.}
        \vspace{-8pt}
        \label{tab:distortion_comparison}
        \begin{tabular}{lccc}
            \toprule
            Model & bpp $\downarrow$ & PSNR $\uparrow$ & MS-SSIM $\uparrow$ \\
            \midrule
            VTM          & 0.0335 & 32.75 & 0.9742 \\
            DCVC-FM      & 0.0328 & 33.29 & 0.9782 \\
            Diff-SIT (ours) & 0.0327 & 28.50 & 0.9387 \\
            \bottomrule
        \end{tabular}

        \vspace{1mm}

        \captionof{table}{Impact of STEM on coding performance. BD-Rates are calculated relative to the configuration without STEM. Negative values indicate bitrate savings.}
        \vspace{-8pt}
        \label{tab:ablation_stem}
        \resizebox{\linewidth}{!}{
            \begin{tabular}{lcc}
                \toprule
                Configuration & BD-Rate (LPIPS) $\downarrow$ & BD-Rate (DISTS) $\downarrow$ \\
                \midrule
                w/o STEM (Anchor) & 0.0\% & 0.0\% \\
                w/ STEM  & -23.6\% & -17.2\% \\
                \bottomrule
            \end{tabular}
        }

    \end{minipage}\hfill
    \begin{minipage}[t]{0.48\linewidth}
        \centering
        \captionof{table}{Ablation study of model components. We compare four configurations We compare four configurations to isolate the contributions of: (1) the One-Step Diffusion (OSD) (2) the sparse encoding strategy (STEM), (3) the Frame Type Embedder (FTE).}
        \label{tab:AblationMain}
        {\renewcommand{\arraystretch}{1.275}
        \begin{tabular}{lccc}
            \toprule
            Configuration & bpp $\downarrow$ & LPIPS $\downarrow$ & DISTS $\downarrow$ \\
            \midrule
            (i) Direct Enc.   & 0.0335 & 0.2319 & 0.1267 \\
            (ii) Direct+OSD & 0.0321 & 0.1863 & 0.1097 \\
            (iii) STEM+OSD  & 0.0317 & 0.1720 & 0.0971 \\
            (iv) \textbf{Ours} & 0.0308 & 0.1689 & 0.0955 \\
            \bottomrule
        \end{tabular}}
        \vspace{4mm}
    \end{minipage}
\end{table}

\noindent \textbf{Quantitative Evaluation.}
As shown in Fig.~\ref{fig:QuantitativeResult}, our proposed Diff-SIT establishes a new state-of-the-art across all evaluated metrics. Our model achieves superior reference perceptual quality, significantly outperforming all baselines on LPIPS and DISTS. Furthermore, Diff-SIT excels in no-reference perceptual quality (CLIPIQA) and temporal consistency (Ewarp), demonstrating a substantial margin over prior methods. Notably, we observe a on-intuitive trend: at ultra-low bitrates, our CLIPIQA and Ewarp scores sometimes become better when the bitrate decreases (e.g., on HEVC Class B). This is attributed to the model synthesizing plausible, realistic textures and ideal motion driven by strong generative priors when explicit conditioning information is insufficient.

It is worth noting that, as shown in Tab.~\ref{tab:distortion_comparison}, our model does not show an advantage in pixel-wise distortion metrics (PSNR, MS-SSIM) compared with traditional methods. This is an expected trade-off. This phenomenon has been discussed in work like~\cite{blau2018perception,blau2019rethinking}, generative compression models are optimized to pursue high perceptual quality at ultra-low bitrates, rather than minimizing pixel-level loss. At these low bitrates, models optimized for distortion (e.g., PSNR) tend to produce overly smooth and blurry images to minimize the mean squared error, which is often inconsistent with human perceptual preferences.

\noindent \textbf{Qualitative Evaluation.}
We present qualitative visual comparisons on three datasets in Fig.~\ref{fig:QualitativeResult}. Our Diff-SIT reconstructions consistently exhibit significantly more fine-grained detail and superior perceptual quality with lower bitrates. In contrast, other methods often suffer from distinct degradation patterns: artifacts, blocks, over-smoothed results with details missing.

\subsection{Ablation Study}

To validate the efficacy of our proposed components, we conduct a comprehensive ablation study. All variants follow an identical training protocol: 700k training steps for Stage 1, followed by 10,000 steps of end-to-end fine-tuning in Stage 2. We compare four configurations: (i) \textbf{Direct Encoding (Baseline)}: a standard baseline where every frame is fully encoded via our backbone compression model. (ii) \textbf{Direct Encoding + One-Step Diffusion}: the baseline output serves as input to the One-Step Diffusion module, which isolate the benefits of generative refinement. (iii) \textbf{STEM + One-Step Diffusion}: use our STEM for compression and then perform a One-Step Diffusion, which can justify the rate-saving capablity of STEM. (iv) \textbf{STEM + ODFTE}: our full model which includes the Sparse Temporal Encoding Module, One-Step Diffusion and Frame Type Embedder (FTE), which shows the effectiveness of FTE.

As shown in Tab.~\ref{tab:AblationMain}, the perceptual quality comparison reveals a clear hierarchy: our full model (iv) outperforms ``(iii) STEM + OSD'', validating Frame Type Embedder's effectiveness in improving quality. (iii)'s superiority over ``(ii) Direct Encoding + OSD'' justify STEM's rate-saving capability. Finally, the substantial performance leap from ``(i) Direct Encoding'' to (ii) serves as compelling evidence for the effectiveness of our one-step video diffusion. More detailed and in-depth ablation analysis for will be provided below.

$ \ $

\noindent \textbf{Effectiveness of Sparse Temporal Encoding (STEM).}
To validate STEM, we benchmark Diff-SIT against the configuration without sparse encoding (Direct Encoding + ODFT) on LPIPS and DISTS. As shown in Tab.~\ref{tab:ablation_stem}, our strategy achieves approximately 20\% bitrate savings (BD-Rate) while maintaining comparable perceptual quality, confirming the high coding efficiency of our sparse representation. This improvement stems from the fact that the low-bitrate MV frames consume negligible bits compared to fully coded backbone frames. Thus, the model can allocate more budget to key structural information when encoding backbone frames, both saving bitrates improving quality.

$ \ $

\begin{figure}[t]
\centering
\includegraphics[width=\linewidth]{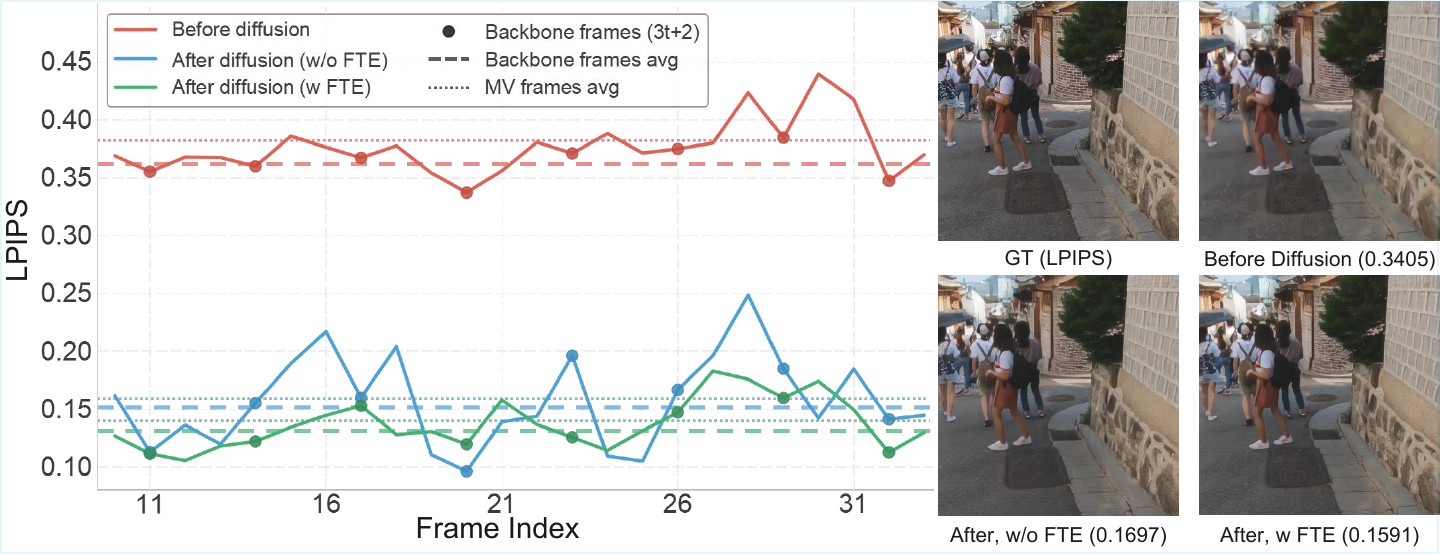}
\caption{Before diffusion, MV frames are initially of much lower quality than backbone frames. (1) However, after diffusion restoration, their perceptual quality becomes much closer to restored P-frames, which indicates ODFTE can exploit information from backbone frames to further improve MV frames' quality. (2) Diffusion with FTE shows better perceptual quality, proving effectiveness of FTE to guide the diffusion process.}
\label{fig:MVafterOFDT}
\end{figure}

\noindent \textbf{Synergy of STEM and ODFTE.}
We analyze the synergy between our sparse encoding (STEM) and generative enhancement (ODFTE). As shown in Fig.~\ref{fig:MVafterOFDT}, (1) the intermediate MV frames (Before Diffusion) initially exhibit significantly lower quality compared to the backbone frames. However, after diffusion restoration, the perceptual quality of the final MV frames becomes much closer to that of the backbone frames. This demonstrates that, despite MV frames' low bitrate, they convey sufficient key information for the diffusion model, which can leverage temporal context for effective restoration. Specifically, the diffusion model generates details and corrects distortions in regions where optical flow estimation fails. This confirms the strong synergy between our sparse encoding (STEM) and generative enhancement (ODFTE). (2) Diffusion model with the Frame Type Embedder (FTE) achieves superior overall quality compared to the variant without it. This validates the benefit of adaptive reconstruction. Conditioned on which type one frame is compressed (I, P or MV), the diffusion model can apply different strength of generation, giving more trust to backbone frames and exploit their information for guiding MV frames' restoration.

$ \ $

\noindent \textbf{Number of MV frames between P frames.}
We vary the number of MV frames between every two P frames. As Tab.~\ref{table: interval} shows, two MV frames (i.e. each backbone frame $\tilde{x}_{3t+2}$ serves as a reference for its two neighboring frames $x_{3t+1}$, $x_{3t+3}$) achieve the best result. It ensures each MV frame is predicted directly from a high-quality backbone frame and maintains low bitrate. Too many MV frames will lead to collapse of optical flow prediction chain, resulting in poor perceptual quality. Meanwhile, too few MV frames will lift the proportion of backbone frames, resulting in high bitrate, and poorer quality if bitrate is fixed. Choosing two MV frames as interval achieves a good trade-off.

\begin{table}[t]
    \centering
    \caption{Ablation on number of MV frames between P frames.}
    \begin{tabular*}{\linewidth}{@{\extracolsep{\fill}}llllll@{}}
        \toprule
        MV frame number & bpp $\downarrow$ & Proportion of backbone frames & LPIPS $\downarrow$ & DISTS $\downarrow$ \\
        \midrule
        Denser: 1 & 0.0355 & 48.5\% (16 backbone / 33 total) & 0.1705 & 0.1012 \\
        Sparser: 4  & 0.0319 & 21.2\% (7 backbone / 33 total) & 0.1823 & 0.1218 \\
        \textbf{Ours: 2} & 0.0308 & 33.3\% (11 backbone / 33 total) & 0.1689 & 0.0955 \\
        \bottomrule
    \end{tabular*}
    \label{table: interval}
\end{table}

$ \ $

\noindent \textbf{Multi-Step Diffusion and Components Latency.}
To quantify the trade-off of our one-step design, we compare our model against multi-step variants and quantify the latency of each component. We implement 2-step and 5-step versions of our ODFTE, keeping all other components (STEM, FTE) identical. As shown in Tab.~\ref{tab:ablation_latency},
\textbf{(1) Diffusion step:} increasing diffusion steps yields a marginal improvement in perceptual quality. However, this comes at a substantial computational cost. The 5-step model, for instance, is approximately 4x slower than our one-step approach. This confirms that our one-step model achieves an excellent trade-off between performance and efficiency, sacrificing minimal perceptual quality for a massive speed-up, making it more practical.
\textbf{(2) Components latency:} the bottleneck is the Wan2.1, resulting in higher decoder latency. The Encoder (STEM) is relatively fast, accounting for 34.8\% of the total latency.

\begin{table}[t]
    \centering
    \caption{Ablation: One-step vs. multi-step diffusion process. We report (1) both the absolute average 
    per-frame inference time (ms) and the relative latency ($\times$) compared to the 1-step baseline. (2) inference latency of each component.}
    \vspace{-4pt}
    \label{tab:ablation_latency}
    \small
    \resizebox{\linewidth}{!}{
        \setlength{\tabcolsep}{3.5mm}
        \begin{tabular}{l c c c}
            \toprule
            (1) Diffuison step & Time (ms / $\times$) $\downarrow$ & LPIPS $\downarrow$ & DISTS $\downarrow$ \\
            \midrule
            ours (1-Step)  & 1275 / 1.00 & 0.1689 & 0.0955 \\
            ours (2-Steps) & 2199 / 1.72 & 0.1615 & 0.0901 \\
            ours (5-Steps) & 4263 / 3.35 & 0.1544 & 0.0832 \\
            \midrule
            Components latency & Wan2.1 & Encoder (STEM) & Decoder (including ODFTE) \\
            \midrule
            Time (ms / percentage) & 656.6 / 51.5\% & 443.7 / 34.8\%  & 831.3 / 65.2\% \\
            \bottomrule
        \end{tabular}
    }
\end{table}

\section{Conclusion}
Traditional video codecs suffer from severe perceptual degradation at ultra-low bitrates, motivating the need for generative reconstruction. In this paper, we proposed Diff-SIT, a novel video compression framework that combines a Sparse Temporal Encoding Module (STEM) with an efficient, One-Step Video Diffusion model with Frame Type Embedder (ODFTE). This integrated approach demonstrates excellent perceptual quality and temporal consistency, especially in the challenging ultra-low-bitrate regime. Experiments confirm that Diff-SIT achieves state-of-the-art performance on multiple benchmarks for video compression, demonstrating its superior efficiency and effectiveness.

\newpage
\bibliographystyle{splncs04}
\bibliography{main}
\end{document}